%% file: 08026-main.tex
\documentclass[runningheads]{llncs}

\usepackage{eccv}

\usepackage{eccvabbrv}

\usepackage{graphicx}
\usepackage{booktabs}

\usepackage{amsmath}
\usepackage{amssymb}

\usepackage{svg}
\usepackage{pifont}

\usepackage{caption}
\usepackage{subcaption}
\usepackage{wrapfig}

\usepackage{multirow,bigdelim}
\usepackage{multicol}
\usepackage{enumitem}

\usepackage{gensymb}
\usepackage{bbm}
\usepackage{bbding} %

\usepackage{lipsum}
\usepackage{makecell}

\usepackage{boldline}
\usepackage{siunitx}
\usepackage{xcolor}

\usepackage{soul}

\usepackage{etoolbox}
\BeforeBeginEnvironment{wraptable}{\setlength{\intextsep}{2pt}}

\definecolor{unknown}{HTML}{FF2400}
\definecolor{background}{HTML}{000000}
\definecolor{bare-ground}{HTML}{F2D8C4}
\definecolor{rocky-terrain}{HTML}{594636}
\definecolor{developed-structures}{HTML}{A6A6A6}
\definecolor{road}{HTML}{52595A}
\definecolor{shrubs}{HTML}{9BE600}
\definecolor{trees}{HTML}{008A35}
\definecolor{sky}{HTML}{00D8F5}
\definecolor{water}{HTML}{0D7FFC}
\definecolor{vehicles}{HTML}{FFF900}
\definecolor{person}{HTML}{FE00AA}

\usepackage[accsupp]{axessibility}  %

\usepackage{hyperref}

\usepackage{orcidlink}

\usepackage{xr-hyper}

\let\llncssubparagraph\subparagraph
\let\subparagraph\paragraph
\usepackage{titlesec}
\let\subparagraph\llncssubparagraph

\titlespacing{\subsubsection}{0pt}{3pt}{3pt}
\titlespacing*{\subsection}{0pt}{6pt}{2pt}
\titlespacing*{\section}{0pt}{8pt}{6pt}

\newcommand\figwidth{0.97}
\newcolumntype{P}[1]{>{\centering\arraybackslash}p{#1}}

\newcommand{\rebuttal}[1]{\textcolor{black}{#1}}

\begin{document}

\title{Caltech Aerial RGB-Thermal Dataset in the Wild}

\author{Connor Lee$^*$ \and
Matthew Anderson$^*$ \and
Nikhil Ranganathan \and
Xingxing Zuo \and \\
Kevin Do \and
Georgia Gkioxari \and
Soon-Jo Chung}

\authorrunning{C. Lee et al.}

\institute{California Institute of Technology \\
\email{\{clee, matta, nrangana, zuox, kdo, georgia, sjchung\}@caltech.edu}}

\maketitle
\def\thefootnote{*}\footnotetext{These authors contributed equally to this work.}
\def\thefootnote{\arabic{footnote}}

\begin{abstract} %
We present the first publicly-available RGB-thermal dataset designed for aerial robotics operating in natural environments. Our dataset captures a variety of terrain across the United States, including rivers, lakes, coastlines, deserts, and forests, and consists of synchronized RGB, thermal, global positioning, and inertial data. We provide semantic segmentation annotations for 10 classes commonly encountered in natural settings in order to drive the development of perception algorithms robust to adverse weather and nighttime conditions. Using this dataset, we propose new and challenging benchmarks for thermal and RGB-thermal (RGB-T) semantic segmentation, RGB-T image translation, and motion tracking. We present extensive results using state-of-the-art methods and highlight the challenges posed by temporal and geographical domain shifts in our data. 
The dataset and accompanying code is available at \url{https://github.com/aerorobotics/caltech-aerial-rgbt-dataset}.

\end{abstract}

\input{sections/intro}

\input{sections/relatedwork}

\input{sections/dataset}
\input{sections/results}

\input{sections/conclusion}

\section*{Acknowledgements}
This work was funded by the Office of Naval Research. We thank K. Koetje, K. Brodie, T. Hesser, T. Almeida, T. Cook, N. Spore, and R. Beach. 

\bibliographystyle{splncs04}

\input{main.bbl}
\end{document}


\title{Caltech Aerial RGB-Thermal Dataset in the Wild Supplementary Material}

\author{Connor Lee$^*$\and
Matthew Anderson$^*$\and
Nikhil Raganathan \and
Xingxing Zuo \and \\
Kevin Do \and
Georgia Gkioxari \and
Soon-Jo Chung}

\authorrunning{C. Lee et al.}

\institute{California Institute of Technology \\
\email{\{clee, matta, nrangana, zuox, kdo, georgia, sjchung\}@caltech.edu}}

\maketitle
\def\thefootnote{*}\footnotetext{These authors contributed equally to this work.}
\def\thefootnote{\arabic{footnote}}

\input{supplemental_material}

\bibliographystyle{splncs04}

\input{08026-supp.bbl}

%% file: sections/intro.tex
\begin{figure}
    \centering
    \includegraphics[width=\linewidth]{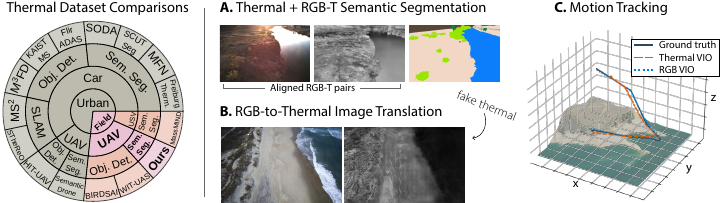}
    \caption{\textit{Left:} Our dataset is uniquely designed to improve thermal scene perception for field robots. \textit{Right:} We provide new benchmarks for thermal-based vision algorithms, including (a) semantic segmentation, (b) image translation, and (c) motion tracking.}
    \label{fig:teaser}
\end{figure}
\section{Introduction}
\label{sec:intro}
Field robots rely predominantly on visual cameras, lidar, and radar to perceive their surroundings~\cite{feng2020deep, fountas2020agricultural}. These sensors provide robust perception capabilities, but falter in low-light and adverse weather conditions~\cite{vargas2021overview}. In contrast, thermal cameras use long-wave infrared wavelengths to capture emitted heat, offering dense, visual information in such conditions where other sensors struggle~\cite{gade2014thermal}. Recently, they have been used for multimodal perception in autonomous driving and are increasingly being used to enable nighttime robot autonomy~\cite{li2020segmenting, XIONG2021103628, vertens2020heatnet, choi2018kaist, yun2022sthereo, nirgudkar2023massmind, deevi2024rgb, li2024riders}. However, successful integration of thermal cameras requires extensive datasets across diverse settings, preventing widespread adoption in field robotics.

Current thermal datasets focus on urban settings for autonomous driving applications (\cref{fig:teaser}). They comprise of thermal-only imagery~\cite{XIONG2021103628, li2020segmenting} or RGB-Thermal (RGB-T) pairs~\cite{flir, ha2017mfnet, vertens2020heatnet, liu2022target}, with some including global positioning (GPS/GNSS) and inertial measurements (IMU) for visual-inertial odometry (VIO) and simultaneous localization and mapping (SLAM)~\cite{yun2022sthereo, choi2018kaist, shin2023deep}. While comprehensive, they lack data depicting natural settings like rivers and forests, which are typical operating areas for field robotics that perform coastline mapping and bathymetry~\cite{brodie_bathy} or activity monitoring during forest fires~\cite{jong2023wit}. 

Due to the lack of thermal benchmarks for algorithms like semantic segmentation and SLAM, field robots cannot easily operate at night or in adverse conditions, requiring online learning~\cite{VS_2022_WACV, lee2023online} or unsupervised domain adaptation to compensate~\cite{gan2023unsupervised, msuda, vertens2020heatnet}, which still require labeled data for evaluation. As such, enabling thermal perception requires collecting and annotating field-specific data from scratch. 
However, creating a thermal dataset for field robotics is difficult due to two reasons:
First, thermal data is hard to crowdsource and web-scrape due to the nicheness of thermal cameras. Hence, acquiring a diverse dataset for algorithm development requires physically going to distinct field locations, increasing time and cost. Second, operating field robots, especially uninhabited aerial vehicles (UAV), usually requires special permits for each location-of-capture. 

In this work, we present the first dataset targeting thermal scene perception for field robotics. The dataset consists of oblique-facing imagery captured from a UAV, supplemented with image sets captured at ground level, and focuses on littoral settings within various desert, forest, and coastal environments across the United States. We contribute new benchmarks for thermal and RGB-T semantic segmentation, RGB-T image translation, and VIO/SLAM algorithms, with unique challenges characterized by temporal and geographical domain shifts for learning-based methods and feature sparsity for motion-tracking algorithms. 
\rebuttal{Aside from the proposed benchmarks, our dataset can also be used towards other applications like visual place recognition and terrain-relative navigation.}

The paper is organized as follows: \cref{sec:related-work} reviews relevant datasets, benchmarks, and algorithms. \cref{sec:dataset} describes our dataset details and curation process, and \cref{sec:results} presents benchmark results. \cref{sec:conclusion} offers concluding remarks.

%% file: sections/relatedwork.tex
\newcolumntype{x}[1]{>{\centering\arraybackslash\hspace{0pt}}p{#1}}

\newcommand{\customcheck}[0]{{\textcolor{green}{\ding{51}}}}
\newcommand{\customcross}[0]{{\textcolor{red}{\ding{55}}}}

\section{Related Work}
\label{sec:related-work}
\subsection{Datasets and Benchmarks}
\subsubsection{Thermal/RGB-T benchmarks for urban robotics}

Current urban thermal and RGB-T datasets primarily focus on autonomous vehicle (AV) technology and surveillance applications. AV-related datasets cover tasks such as object detection~\cite{flir, liu2022target, choi2018kaist}, semantic segmentation~\cite{li2020segmenting, ha2017mfnet, vertens2020heatnet, XIONG2021103628}, and localization~\cite{yun2022sthereo, choi2018kaist, shin2023deep}. Some include data from other spectra, benchmarking not only thermal algorithms but also RGB-T~\cite{liu2022target, choi2018kaist, ha2017mfnet, vertens2020heatnet, speth2022deep} and other multispectral models. In contrast, datasets for surveillance applications use fixed cameras~\cite{jia2021llvip} and UAVs~\cite{suo2022hit, speth2022deep} to detect personnel, vehicles, and other objects. Despite many urban thermal benchmarks, they cannot be directly used to develop algorithms for robots operating in natural environments due to the domain gap.

\subsubsection{Thermal/RGB-T benchmarks for field robotics} 
Datasets for robots operating in non-urban environments remain limited. MassMIND~\cite{nirgudkar2023massmind} and PST900~\cite{shivakumar2020pst900} benchmark thermal semantic segmentation for maritime and subterranean robotic environments, respectively. WIT-UAS~\cite{jong2023wit} provides UAV-borne thermal images for object detection in wildfire-prone environments. One work~\cite{lee2023online} proposes an aerial RGB-T dataset for thermal water segmentation in littoral areas but only provides 1272 labels consisting of a single \textit{water} class. Benchmarks like \cite{bondi2020birdsai, noaaseals} depict natural environments but are targeted towards wildlife tracking. To address this gap, we present an aerial RGB-T dataset which can be used to benchmark multiple tasks, including semantic segmentation, RGB-T image translation, and localization, across diverse natural environment robotic applications. Specifically, we build upon the dataset from~\cite{lee2023online} by extending the number of semantic classes from 1 to 10 and adding additional RGB-T data captured from air and ground, resulting in a total of 4195 semantic segmentation annotations. \rebuttal{In comparison to existing thermal/RGB-T datasets, our dataset can enable nighttime field robotic applications like remote sensing research, geological sampling, precision agriculture, and wilderness search and rescue.}

\subsubsection{UAV-based semantic segmentation datasets}
Aside from~\cite{lee2023online}, existing UAV-captured semantic segmentation datasets~\cite{nigam2018ensemble, chen2018large, LYU2020108, cai2023vdd, semanticDroneDataset, speth2022deep} cover urban environments (\cref{tab:related-datasets}). With the exception of~\cite{semanticDroneDataset} which provides 400 thermal image samples, all of these datasets contain only RGB data \rebuttal{for semantic segmentation}. While they provide images with relevant view angles, the modality gap and focus on urban content makes them impractical for aerial field robotic applications. 

\begin{table*}[t]
    \setlength{\tabcolsep}{10pt}
    \centering
    \caption{Comparison of datasets captured from aerial platforms. Camera angle of 90\degree is nadir-pointing. See Tab. S1 in supplement for comparison against thermal datasets captured from surface and ground vehicles.}
    \resizebox{\linewidth}{!}{%
    \begin{tabular}{llccccllcc}
        \toprule[1pt]
        Dataset & Task & RGB & Thermal & GPS & IMU & Setting & Location & \# Samples & \makecell{Camera pose} \\
        \midrule[1pt]

        VisDrone~\cite{zhu2021detection}  & Obj. Det. & \customcheck & \customcross  & \customcross  & \customcross & Urban & China & 10,209  & Variable \\
        \midrule[0.25pt]
        
        HIT-UAV~\cite{suo2022hit}, NII‐CU~\cite{speth2022deep}  & Obj. Det. & \customcheck & \customcheck & \customcross & \customcross & Urban & China, Japan & 2,898 - 5,880 & 30\degree-90\degree, 60-130 m \\
        \midrule[0.25pt]
        
        WIT-UAS~\cite{jong2023wit}  & Obj. Det. & \customcross & \customcheck & \customcheck & \customcheck & \makecell[l]{Forest} & USA & 6,951 & 30\degree, 90\degree, < 120 m \\
        \midrule[0.25pt]

        BIRDSAI~\cite{bondi2020birdsai}  & Obj. Track. & \customcross & \customcheck & \customcross & \customcross & Savannas & S. Africa & 62,000 & Off-nadir, 60-120 m \\
        \midrule[0.25pt]
        
        \makecell[l]{AeroScapes~\cite{nigam2018ensemble}, UAVid~\cite{LYU2020108},\\~\cite{chen2018large},~\cite{cai2023vdd}, Swiss/Okutama~\cite{speth2022deep}} 
         & Sem. Seg. & \customcheck & \customcross & \customcross  & \customcross & Urban & \makecell[l]{China, Japan, \\ Switzerland} & 191 - 3269 & 30 - 90\degree, < 100 m \\
        \midrule[0.25pt]
        
        \makecell[l]{Semantic Drone Dataset~\cite{semanticDroneDataset}}  & Sem. Seg. & \customcheck & \customcheck & \customcross & \customcheck & Urban & Germany & 400 & 90\degree, < 30m \\
        \midrule[0.25pt]

        \textbf{Ours} & Sem. Seg. & \customcheck & \customcheck & \customcheck & \customcheck & \makecell[l]{River, Coast, Lake, \\Mountain, Desert} & USA & 4,195 & 20\degree, 45\degree, < 120 m \\

        \bottomrule[1pt]
    \end{tabular}
    }%
    \label{tab:related-datasets}
\end{table*}

\subsection{Algorithms using Thermal Datasets}
\subsubsection{Thermal semantic segmentation}
Since thermal images are spatially identical to their RGB counterparts, popular RGB semantic segmentation algorithms~\cite{poudel2019fast, chen2018encoder, cai2022efficientvit, xie2021segformer} can be applied directly. However, some works~\cite{li2020segmenting, XIONG2021103628, ftnet} develop models specifically for thermal imagery, and leverage edge priors to overcome low resolution and blurriness due to thermal crossover~\cite{retief2003prediction}.  
Although specifically designed for thermal, these models only marginally outperform standard RGB segmentation models~\cite{ftnet} and their performance in field settings, where edges are less structured, remains untested. Aside from architectural advancements, improvements in thermal semantic segmentation come from domain adaptation (DA) methods~\cite{gan2023unsupervised, li2020segmenting, vertens2020heatnet, Ustun_2023_CVPR} that use labeled RGB data with unlabeled thermal data. To provide fair comparisons, we do not provide DA baselines in our benchmarks as performance varies based on the choice of RGB data.

\subsubsection{RGB-T semantic segmentation}
RGB-T semantic segmentation methods compensate for the weaknesses of each modality via deep feature fusion, data augmentation, and adversarial training. Most approaches~\cite{ha2017mfnet, fuseseg, RTFNet, shin2023complementary, zhang2023delivering} utilize encoder-decoder architectures with modality-specific encoders and shared decoders, and show significant improvements over channel-stacked RGB-T inputs. Works like~\cite{eaefnet, gmnet} make improvements by integrating RGB and thermal features at multiple stages within the encoders, while \cite{shivakumar2020pst900} eschews RGB-T training pairs by processing RGB-T inputs sequentially. Other methods include random input masking to reduce reliance on a single modality~\cite{shin2023complementary} and adversarial training to adapt to different times of day~\cite{vertens2020heatnet}.

\subsubsection{RGB-T image translation}
RGB-T image translation is vital as it may enable thermal training data to be generated from large-scale RGB datasets. Works based on generative adversarial networks (GAN) such as~\cite{unit, munit, zhu2017unpaired, park2020contrastive, lee2023edge} can be used without RGB-T image pairs. However, GANs like~\cite{isola2017image, wang2018pix2pixHD} that require paired imagery provide better translations for both visual appeal and domain adaptation~\cite{li2020segmenting}. Recent diffusion methods~\cite{saharia2022palette, Rombach_2022_CVPR} have demonstrated remarkable results on RGB imagery, but have not been tested on RGB-T image translation.

\subsubsection{Thermal VIO and SLAM}
Low signal-to-noise ratio, reduced contrast, and blurred boundaries in thermal images present challenges for VIO and SLAM. Several studies~\cite{bloesch2017iterated, delaune2019thermal, he2021using, hua2021i2, doer2021radar} use indirect methods by tracking keypoints extracted from thermal images for visual constraints and rely on image normalization techniques to improve keypoint extraction. In contrast,~\cite{khattak2019keyframe} uses a direct method to extract and track high-gradient points by minimizing radiometric error. Other works utilize learned local and global features to improve feature tracking~\cite{zhao2020tp} and loop closure detection~\cite{saputra2021graph}, respectively.

%% file: sections/dataset.tex
\section{The Caltech Aerial RGB-Thermal Dataset}
\label{sec:dataset}

\subsection{Data Acquisition Hardware}
\begin{figure}
    \centering
    \includegraphics[width=\figwidth\textwidth]{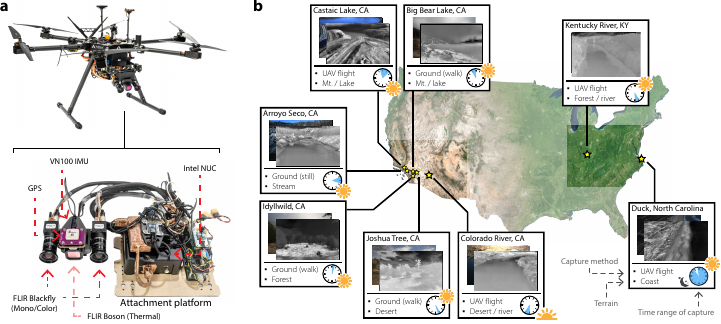}
    \caption{\textbf{(a)} The Aurelia X6 hexacopter and sensor stack used to capture our aerial dataset. 
    \textbf{(b)} Geographic distribution of our data collection sites and collection times.}
    \label{fig:schematic}
\end{figure}

To capture this dataset, we developed a custom sensor stack with synchronized RGB-thermal imagery, IMU, and global positioning data (\cref{fig:schematic}). The sensor stack features a non-radiometric FLIR ADK thermal camera (640$\times$512~px and 75\textdegree~horizontal FoV, \qty{60}{Hz}) flanked by a pair of FLIR Blackfly electro-optical (EO) monochrome and color cameras\footnote{BFLY-U3-23S6M-C mono/color with Tamron M112FM08 lenses} (960$\times$600~px and 75\textdegree~horizontal FoV, \qty{30}{Hz}).
Pose information is provided by a VectorNav VN100 IMU (\qty{200}{Hz}) and a u-blox M8N GPS (\qty{5}{Hz}). 
All three cameras and the VN100 are hardware synchronized using a dedicated signal generator and are rigidly attached to each other using a stiff, 3D printed mount.
A Simply NUC Ruby Mini PC (AMD Ryzen 7 4800U, 32~GB RAM) is used for the compute, and the entire system is interfaced using ROS Noetic.
Our cameras and IMU were calibrated following the procedures found in Supplementary Material Sec.~S2.2 and were not disassembled from each other between any of the dataset collections.

The sensor stack is mounted rigidly onboard an Aurelia X6 hexacopter (\cref{fig:schematic}) without gimbal stabilization at angles between 0\textdegree~(level with the horizon) to 45\textdegree~downwards to provide different viewpoints. 
When mounted to a UAV, we also log the position and attitude estimates from the UAV at \qty{20}{Hz} providing improved positioning accuracy from the onboard RTK GPS.
Additionally, the sensor stack's modularity allows us to capture datasets on foot by mounting it to a tripod where flight restrictions are in-place.

\subsection{Data Collection and Processing}

\subsubsection{Data capture}
We captured 37 aerial and ground trajectories covering lakes, rivers, coastlines, deserts, and mountains from around Southern California, Kentucky, and North Carolina (\cref{fig:schematic}). Aerial trajectories (18) involve high motion with intermittent hovering, and mostly depict scenes from \qty{40}{m} altitude. Ground-level trajectories (19) were captured on foot, with 7 captured without any movement for VIO debugging purposes. For more details, see Supplementary Materials Tab.~S2 and Sec.~S2.1.

\subsubsection{Thermal image normalization}
\label{section:thermal16bit-normalization}
For the baselines presented in this work, we typically normalize 16-bit thermal data by rescaling between the 1\textsuperscript{st} and 99\textsuperscript{th} percentile pixel values and follow with contrast limited adaptive histogram equalization (CLAHE). We keep normalized data as floats to minimize information loss, and only convert to 8-bit format when necessary or for visualization. 
\begin{figure}[htbp]
    \centering
    \includegraphics[width=\figwidth\textwidth]{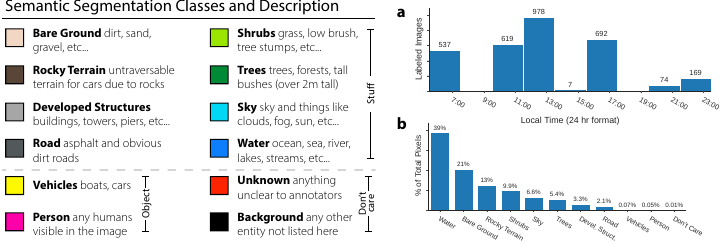}
    \caption{Semantic segmentation classes in our dataset. The color mapping is used throughout this paper. \textbf{(a)} Hourly distribution of annotated thermal images. \textbf{(b)} Histogram of semantic classes.} 
    \label{fig:segmentation-class}
\end{figure}
\subsubsection{Semantic segmentation annotation}
\label{sec:dataset-semseg-annotation}
To aid development of scene perception algorithms for field robots, we manually-annotated a subset of thermal images for semantic segmentation with classes in~\cref{fig:segmentation-class}. To avoid redundancy, images were sampled every \qty{3}{s} when moving and every \qty{20}{s} when still, resulting in 4195 samples. We considered a frame in motion if the distance moved within the past 2 seconds exceeded 0.5 meters, with distances computed using GPS coordinates. 

\rebuttal{To promote labeling consistency, annotators preread descriptions of each trajectory, which contained a list of classes present, labeled examples provided by the authors, and GPS coordinates (for Google Earth look-up). Annotators were then given an RGB-T image pair with trajectory metadata and asked to label the thermal frame. Annotations were reviewed by authors, with inconsistent or incorrect labels sent back. 3 rounds of annotation were done, with the final and initial set of labels differing by 0.08 mIoU. Classes most frequently re-annotated were \textit{bare ground}, \textit{rocky terrain}, \textit{road}, \textit{trees}, and \textit{vehicles}.}

\subsubsection{RGB-thermal image alignment}
\label{sec:dataset-rgbt-alignment}
To align thermal and RGB image pairs, we stereo rectified the RGB-T image pairs using the camera matrices obtained from calibration (Sec.~S2.2) and projected the thermal image into the larger RGB image frame to preserve RGB resolution. As our baseline is small compared to the depth of the scenes we capture, image pairs are near-coregistered after rectification. Due to differences in optics, the field-of-view of the aligned image pairs is narrower compared to that of the thermal camera.

\subsection{Dataset Splits}
\subsubsection{General (benchmark) split}\label{sec:dataset-general-split}
This is our primary split for semantic segmentation and image translation benchmarks. We randomly partition the annotated thermal dataset (\cref{sec:dataset-semseg-annotation}) into train/val/test sets at a 75:12.5:12.5 ratio. 

\subsubsection{Temporal split}\label{sec:dataset-temporal-split}
To promote studies into the effect of different time-of-day capture of the thermal images, we split the dataset using three time periods: twilight/sunrise (5\,AM - 7\,AM), daytime (10\,AM - 5\,PM), and nighttime (7\,PM - 4\,AM). We randomly partition the daytime images into train/val/test sets using the ratio from above and leave the twilight and nighttime sets for testing.  

\subsubsection{Geographical split}\label{sec:dataset-geographic-split}
The geographical splits are intended to test algorithm adaptability to unseen settings. We provide two splits:
\begin{enumerate}
    \item \textit{Terrain-based:} This split partitions data into the following terrain categories: aerial river, aerial lake, aerial coast, and ground-captured images. Train/val/test splits are created per category with a 50:15:35 ratio.
    \item \textit{Region-based:} This is based on the area of the United States where the data was captured. Distinct regions include California (CA), Kentucky (KY), and North Carolina (NC), with train/val/test created in the same way as above.
\end{enumerate}

%% file: sections/results.tex
\section{Experiments}
\label{sec:results}
We provide baselines for the following benchmarks: thermal semantic segmentation, RGB-T semantic segmentation, RGB-T image translation, and VIO/SLAM. Furthermore, we explore the zero-shot thermal segmentation capabilities of large vision-language models on our dataset and perform further ablation studies.

\begin{table}[h]
\setlength{\aboverulesep}{2pt}
    \setlength{\belowrulesep}{2pt}
        \centering
	\caption{Supervised model performance (mIoU), framerate (FPS at batch size 1), and floating point operations (FLOP) for various semantic segmentation network baselines.}
	\label{tab:supervised-baselines}
	\resizebox{\linewidth}{!}{%
		\begin{tabular}{p{4cm}P{1.25cm}P{1.25cm}P{1.25cm}P{1.25cm}P{1.25cm}P{1.25cm}P{1.25cm}P{1.25cm}P{1.25cm}P{1.25cm}P{1.05cm}P{1.05cm}P{1.05cm}P{1.15cm}P{1.15cm}P{1.15cm}}
\toprule[1pt]

\makecell[l]{Model} & \rotatebox[origin=c]{0}{\makecell{Bare\\ ground}} &  \rotatebox[origin=c]{0}{\makecell{Rocky \\ terrain}} &  \rotatebox[origin=c]{0}{\makecell{Develop. \\ struct.}} &  \rotatebox[origin=c]{0}{\makecell{Road}} &  \rotatebox[origin=c]{0}{\makecell{Shrubs}} &  \rotatebox[origin=c]{0}{\makecell{Trees}} &  \rotatebox[origin=c]{0}{\makecell{Sky}} &  \rotatebox[origin=c]{0}{\makecell{Water}} &  \rotatebox[origin=c]{0}{\makecell{Vehicles}} &  \rotatebox[origin=c]{0}{\makecell{Person}} &  \makecell{Stuff} & {\makecell{Object}} & {\makecell{All}} & \makecell{Orin \\ (GPU)} & \makecell{NUC \\(CPU)} & \makecell{FLOPS \\ (G)}  \\ \midrule[1pt]

FastSCNN~\cite{poudel2019fast} & 0.807 & 0.907 & 0.728 & 0.631 & 0.694 & 0.776 & 0.956 & 0.965 & 0.384 & 0.058 & 0.808 & 0.221 & 0.690 & \textbf{124.2} & \textbf{85.6} & \textbf{1.0} \\
MobileNetV3-S 0.75~\cite{howard2019searching} & 0.802 & 0.914 & 0.676 & 0.633 & 0.704 & 0.814 & 0.955 & 0.976 & 0.383 & 0.111 & 0.809 & 0.247 & 0.697 & 75.2 & 24.0 & 3.3 \\
MobileNetV3-L 0.75 & 0.801 & 0.914 & 0.701 & 0.615 & 0.713 & 0.793 & 0.959 & \textbf{0.976} & 0.432 & 0.148 & 0.809 & 0.290 & 0.705 & 59.0 & 15.5 & 4.8 \\
MobileViTV2 0.50~\cite{mehta2022separable} & 0.822 & 0.913 & 0.759 & 0.650 & 0.694 & 0.781 & 0.954 & 0.970 & 0.222 & 0.030 & 0.818 & 0.126 & 0.680 & 60.1 & 10.7 & 5.5 \\
EfficientViT-B0~\cite{cai2022efficientvit} & 0.825 & 0.917 & \textbf{0.777} & 0.618 & 0.710 & 0.793 & 0.961 & 0.965 & 0.495 & \textbf{0.191} & 0.821 & \textbf{0.343} & \textbf{0.725} & 51.2 & 5.5 & 3.9 \\
EfficientNet-Lite0~\cite{tan2019efficientnet} & 0.819 & 0.908 & 0.769 & 0.601 & 0.709 & 0.805 & 0.962 & 0.968 & 0.380 & 0.079 & 0.818 & 0.229 & 0.700 & 47.7 & 12.8 & 7.2 \\
EfficientNet-Lite2 & \textbf{0.825} & 0.920 & 0.765 & 0.614 & 0.698 & 0.805 & 0.958 & 0.971 & 0.386 & 0.118 & 0.820 & 0.252 & 0.706 & 38.9 & 11.2 & 9.4 \\
Segformer-B0~\cite{xie2021segformer} & 0.804 & 0.910 & 0.690 & 0.624 & 0.696 & 0.788 & 0.960 & 0.970 & 0.195 & 0.000 & 0.805 & 0.097 & 0.664 & 38.3 & 5.9 & 10.2 \\
Segformer-B1 & 0.814 & 0.909 & 0.773 & 0.603 & 0.713 & 0.799 & 0.960 & 0.965 & 0.366 & 0.000 & 0.817 & 0.183 & 0.690 & 30.3 & 3.5 & 19.9 \\
ResNet18~\cite{he2016deep} & 0.810 & 0.905 & 0.766 & 0.618 & 0.711 & 0.807 & 0.959 & 0.966 & 0.431 & 0.158 & 0.818 & 0.294 & 0.713 & 69.3  & 12.4 & 22.4\\
ResNet50 & 0.819 & 0.916 & 0.747 & \textbf{0.658} & 0.711 & 0.799 & 0.961 & 0.974 & 0.361 & 0.182 & 0.823 & 0.272 & 0.713 & 29.9  & 6.1  & 45.4\\
ResNeXt50~\cite{xie2017aggregated} & 0.825 & 0.919 & 0.773 & 0.653 & 0.709 & 0.794 & 0.964 & 0.976 & 0.436 & 0.137 & \textbf{0.827} & 0.287 & 0.719 & 23.0 & 5.6 & 45.5 \\
ConvNext-T~\cite{liu2022convnet} & 0.799 & 0.909 & 0.719 & 0.608 & 0.706 & 0.808 & 0.961 & 0.970 & 0.363 & 0.003 & 0.810 & 0.183 & 0.685 & 25.8 & 3.3 & 47.4 \\
ConvNext-S & 0.810 & 0.913 & 0.771 & 0.603 & \textbf{0.718} & 0.810 & 0.965 & 0.964 & 0.492 & 0.048 & 0.819 & 0.270 & 0.709 & 19.1 & 2.3 & 75.0 \\
ConvNext-B & 0.810 & \textbf{0.921} & 0.697 & 0.611 & 0.706 & 0.803 & 0.963 & 0.972 & 0.348 & 0.112 & 0.811 & 0.230 & 0.694 & 14.6 & 1.4 & 130.5 \\
ConvNext-B (CLIP)~\cite{radford2021learning} & 0.813 & 0.918 & 0.683 & 0.632 & 0.718 & 0.813 & \textbf{0.965} & 0.972 & \textbf{0.517} & 0.137 & 0.814 & 0.327 & 0.717 & 14.6  & 1.4  & 130.4\\
ConvNext-B (CLIP)~\SnowflakeChevron & 0.773 & 0.887 & 0.713 & 0.487 & 0.656 & 0.756 & 0.946 & 0.951 & 0.158 & 0.019 & 0.771 & 0.089 & 0.635 & 14.6  & 1.4  & 130.4\\
DINOv2~\cite{oquab2023dinov2} (linear head) & 0.800 & 0.907 & 0.681 & 0.606 & 0.693 & 0.796 & 0.956 & 0.967 & 0.375 & 0.079 & 0.801 & 0.227 & 0.686 & 15.2 & 2.1 & --- \\
DINOv2 (linear head)~\SnowflakeChevron & 0.705 & 0.844 & 0.619 & 0.419 & 0.558 & 0.725 & 0.922 & 0.920 & 0.356 & 0.158 & 0.714 & 0.257 & 0.623 & 15.2 & 2.1 & --- \\
DINOv2 (nonlin. head) & 0.810 & 0.916 & 0.708 & 0.635 & 0.700 & \textbf{0.819} & 0.959 & 0.973 & 0.399 & 0.119 & 0.815 & 0.259 & 0.704 & 15.2 & 2.1 & --- \\
DINOv2 (nonlin. head)~\SnowflakeChevron & 0.796 & 0.903 & 0.732 & 0.606 & 0.691 & 0.794 & 0.954 & 0.963 & 0.471 & 0.149 & 0.805 & 0.310 & 0.706 & 15.2 & 2.1 & --- \\
FTNet$^\dagger$~\cite{ftnet}            & 0.755 & 0.867 & 0.635 & 0.576 & 0.643 & 0.653 & 0.787 & 0.947 & 0.234 & 0.024 & 0.733 & 0.129 & 0.613 & 11.1  & 1.4  & 100.5\\
\bottomrule[1pt]
		\end{tabular}
	}
	\vspace{-0.1cm}
	\captionsetup{font=scriptsize}
	\caption*{$^\dagger$Thermal-specific model \quad \SnowflakeChevron~Frozen encoder}
\end{table}

\subsection{Thermal Semantic Segmentation}
\label{sec:thermal-sem-seg}
We run multiple baselines on our general split before choosing a specific model to test on the geographic and temporal splits. We go over our baselines before analyzing their performance on our benchmarks. 

\subsubsection{Baselines}
Motivated by robotic applications, we are interested in both inference speed and segmentation performance. Our baselines place an emphasis on lightweight, real-time networks, but also on foundation models like DINOv2~\cite{oquab2023dinov2} and ConvNext-B (CLIP)~\cite{liu2022convnet, radford2021learning} to explore any latent multimodal capabilities. We also test a thermal-specific model, FTNet~\cite{ftnet}, which uses an edge-based loss function to compensate for blurred boundaries in thermal images.  

Besides FastSCNN~\cite{poudel2019fast}, EfficientViT~\cite{cai2022efficientvit}, Segformer~\cite{xie2021segformer}, DINOv2, and FTNet, all network encoders in \cref{tab:supervised-baselines} feed into a DeepLabV3+ segmentation head. In particular, DINOv2 was employed with linear and non-linear, multi-scale heads in order to study the generalization capacity of its pretrained features to the thermal modality. All baselines except DINOv2 and ConvNext-B (CLIP) were trained from pretrained ImageNet weights. All networks, besides FTNet, were trained using the cross entropy loss. Further details on baseline implementation and training can be found in Supplementary Material Sec.~S3.1.

\begin{figure}
    \centering
    \includegraphics[width=\figwidth\linewidth]{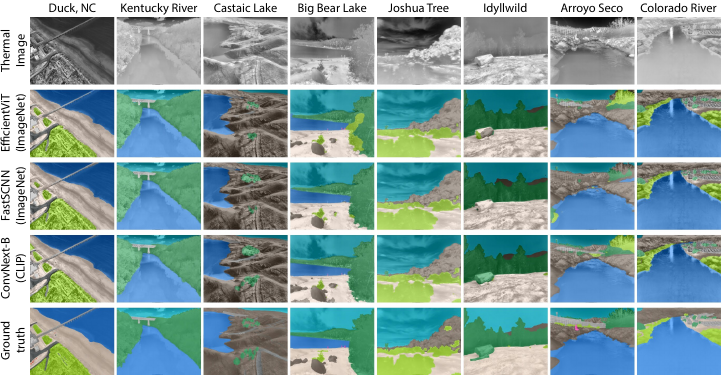}
    \caption{Thermal images and semantic segmentation labels from each capture area with inference results from EfficientViT, FastSCNN, and ConvNext-B (CLIP).}
    \label{fig:thermal-segmentation-and-dataset-examples}
\end{figure}

\subsubsection{General benchmark}
\label{sec:random-split}
We use the general (random) split (\cref{sec:dataset-general-split}) to compare thermal semantic segmentation performance between baselines (\cref{tab:supervised-baselines}). With field robotic applications in mind, we analyze segmentation results in context of GPU and CPU frame rates, which were measured onboard the Nvidia Jetson AGX Orin (GPU) and the Ruby NUC (CPU) embedded platforms, respectively\footnote{Models of input size 512$\times$640 were converted to ONNX and inference speed was averaged over 50 forward passes after a warm-up period.}. 

In general, we found that larger models provide marginal benefit over smaller, compute-efficient models. In particular, a 0.027 mIoU difference between the largest (ConvNext-B) and fastest (FastSCNN) baselines comes at the cost of 130$\times$ more FLOPs. Overall, EfficientViT-B0 performed the best, attaining the highest mIoU (0.725) while performing \qty{50}{Hz} inference on the Jetson AGX Orin. 

While large RGB foundation models show little gain over ImageNet pretraining, their features do extend to the thermal modality. Specifically, a frozen DINOv2 (linear head) outperforms the thermal-specific FT-Net. With a nonlinear head, it matches the performance of other end-to-end trained networks. While this shows the benefit of large-scale RGB pretraining, the choice of RGB pretraining data or learning strategy seems to matter. Notably, the frozen ConvNext-B (CLIP) model gives marginal gain over the frozen DINOv2 (linear head), despite a larger, nonlinear segmentation head. Overall, foundation models do not confer significant performance advantages for semantic segmentation on our dataset in order to justify their computational cost.

Overall, all baselines can segment \textit{stuff} classes well but none excel with the rare \textit{object} classes: \textit{vehicle} and \textit{person}. Future models would likely need to consider weighted loss functions, smarter data augmentation, or domain adaptation techniques in order to improve on object classes. Lastly, we note that the thermal-specific model, FTNet, performs poorly, likely as its edge priors are less effective in unstructured, natural environments compared to urban environments.

\begin{figure}[h]
    \centering
    \includegraphics[width=\figwidth\linewidth]{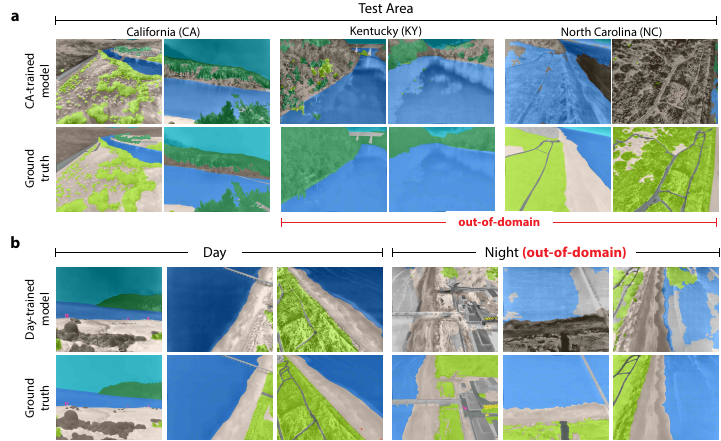}
    \caption{(a) Thermal semantic segmentation failures due to intra-class semantic variations when testing on geographically-partitioned, \textit{out-of-domain }data. (b) Failures due to photometric variations between day \textit{(\textit{in-domain}) }and night \textit{(out-of-domain)}.}
    \label{fig:failure-cases-benchmarks}
\end{figure}

\subsubsection{Geographically-partitioned benchmarks} 
We retrained our best baseline\\(Tab.~\ref{tab:supervised-baselines}) on geographically-partitioned sets to evaluate intra-thermal generalization. Given the cost and effort of collecting diverse aerial thermal data of field settings, we aim to determine the minimum amount of annotated data required to effectively deploy. Using EfficientViT-B0, we performed two experiments to study generalization across difference geographic regions and terrain types.
\begin{table}[htbp]
        \centering
	\caption{Thermal semantic segmentation results on geographically-split data}
	\vspace{-0.25cm}
	\begin{subtable}[t]{0.45\linewidth}
		\setlength{\aboverulesep}{1pt}
		\setlength{\belowrulesep}{1pt}
		\centering
		\caption{Region-based split}
		\label{tab:area-split}
		\resizebox{\linewidth}{!}{%
			\begin{tabular}{p{2cm}P{1.15cm}P{1.15cm}P{1.15cm}P{1.15cm}}
				\toprule[1pt]
				\multirow{2}{*}[-2pt]{Train Area} & \multicolumn{3}{c}{Test Area} & \multirow{2}{*}[-2pt]{\makecell{Avg.\\ mIoU}} \\
				\cmidrule(lr){2-4}
				    & CA    & NC    & KY    &\\
				\midrule[1pt]
				CA  & 0.666 & 0.084 & 0.193 & 0.314\\
				NC  & 0.084 & 0.508 & 0.031 & 0.208\\
				All & 0.648 & 0.525 & 0.587 & 0.586\\
				\bottomrule[1pt]
			\end{tabular}
		}
	\end{subtable}
	\hspace{0.1cm}
	\begin{subtable}[t]{0.45\linewidth}
		\setlength{\aboverulesep}{1pt}
		\setlength{\belowrulesep}{1pt}
		\centering
		\caption{Terrain-based split}
		\label{tab:terrain-split}
		\resizebox{\linewidth}{!}{%
			\begin{tabular}{p{3cm}P{1.15cm}P{1.15cm}P{1.15cm}P{1.15cm}}
				\toprule[1pt]
				\multirow{2}{*}[-2pt]{Training Terrain} & \multicolumn{3}{c}{Testing Terrain} & \multirow{2}{*}[-2pt]{\makecell{Avg.\\ mIoU}} \\
				\cmidrule(lr){2-4}
				             & River & Lake  & Coast &\\
				\midrule[1pt]
				River        & 0.668 & 0.196 & 0.117 & 0.327\\
				Lake         & 0.196 & 0.364 & 0.061 & 0.207\\
				Coast        & 0.127 & 0.119 & 0.522 & 0.256\\
				All          & 0.646 & 0.415 & 0.493 & 0.518\\
				All + ground & 0.658 & 0.416 & 0.520 & 0.532\\
				\bottomrule[1pt]
			\end{tabular}
		}
	\end{subtable}
\end{table}

In the first experiment, we trained on thermal data from CA and tested on images from KY and NC (\cref{tab:area-split}). Despite common classes in each region, results show poor generalization to class variations across geographic areas. We see this again when repeating with the NC split. In the second experiment, we divided our data into three terrain types, \textit{aerial river}, \textit{aerial lake}, and \textit{aerial coast}. Remaining ground-captured images were lumped into a fourth \textit{ground} category. We trained and tested all pairwise combinations and found poor generalization across terrain types as well (\cref{tab:terrain-split}). Notably, we found low, in-domain performance for the lake setting but saw gains after training on aerial data from other terrain and further improvements after training on ground level images. Overall, our results reiterate the benefit of having training data representative of the testing environment, but also show that gains can be made by including out-of-domain (OOD) data from different terrain and different viewpoints.

Overall, the geographic benchmark presents a difficult domain adaptation problem, requiring both inter- and intra-modality domain adaptation techniques to overcome. Although such techniques could perform well in this challenge, the results also highlight a need for an even larger and more diverse thermal dataset capturing common field environments than what we presented.

\subsubsection{Temporally-partitioned benchmark}

Here, we assess EfficientViT-B0's performance against intra-thermal, temporal distribution shift due to thermal inversion. We train the model on images from the daytime and test on images captured at sunrise and at night. The model achieved mIoUs of 0.242 at sunrise, 0.193 at night, and 0.777 on a separate daytime test split. Despite training on daytime images from the same location, the model fails on nighttime scenes, struggling to classify \textit{water} due to thermal inversion (\cref{fig:failure-cases-benchmarks}b). As such, to perform well on this split, future algorithms need to augment datasets to simulate such inversions or intentional collect data capturing such phenomena.  

\begin{table}[htbp]
	\setlength{\aboverulesep}{1pt}
	\setlength{\belowrulesep}{1pt}
	\centering
	\caption{RGB-Thermal semantic segmentation network baseline results (mIoU).}
	\label{tab:rgbt-supervised-baselines}
	\resizebox{\linewidth}{!}{%
		\begin{tabular}{p{3.5cm}P{1.15cm}P{1.15cm}P{1.15cm}P{1.15cm}P{1.15cm}P{1.15cm}P{1.15cm}P{1.15cm}P{1.15cm}P{1.15cm}P{1.15cm}P{1.25cm}}
			\toprule[1pt]

			\makecell{Model} & \makecell{Bare\\ ground} & \makecell{Rocky \\ terrain} & \makecell{Devel. \\ struct.} &  \makecell{Road} &  \makecell{Shrubs} &  \makecell{Trees} &  \makecell{Sky} &  \makecell{Water} &  \makecell{Vehicles} &  \makecell{Person} &  \makecell{All} & \makecell{FLOPS \\ (G)} \\
			\midrule[1pt]
			RGB Only$^\dagger$                & 0.790 & 0.873 & 0.808 & 0.559 & 0.664 & 0.752 & 0.891 & 0.967 & 0.399 & 0.000 & 0.670 & 45\\
			Thermal Only$^\dagger$            & 0.766 & 0.835 & 0.789 & 0.536 & 0.654 & 0.749 & 0.907 & 0.971 & 0.399 & 0.000 & 0.661 & 45\\
                EAEFNet~\cite{eaefnet}            & 0.805 & 0.863 & 0.813 & 0.545 & 0.702 & 0.802 & \textbf{0.937} & 0.976 & 0.481 & 0.000 & 0.692 & 358 \\
			CRM~\cite{shin2023complementary}  & 0.816 & \textbf{0.900} & 0.851 & \textbf{0.616} & 0.695 & 0.779 & 0.923 & 0.974 & 0.377 & \textbf{0.206} & 0.714 & 310\\
			CMNeXt~\cite{zhang2023delivering} & \textbf{0.830} & \textbf{0.900} & \textbf{0.861} & 0.607 & \textbf{0.718} & \textbf{0.808} & 0.933 & \textbf{0.980} & \textbf{0.560} & 0.190 & \textbf{0.740} & 148\\
			\bottomrule[1pt]
		\end{tabular}
	}
	\vspace{-0.1cm}
	\captionsetup{font=scriptsize}
	\caption*{$^\dagger$ DeepLabV3+ (w/ ResNet50 encoder)}
\end{table}
\subsection{RGB-T Semantic Segmentation}
\label{sec:rgbt-sem-seg}
We use the paired RGB-T dataset (\cref{sec:dataset-rgbt-alignment}) and partition it using the general split (\cref{sec:random-split}). Recall that RGB and thermal input sizes are 960$\times$600 pixels and the thermal FoV is narrower than in the previous benchmarks (\cref{sec:thermal-sem-seg}).

\subsubsection{Baselines}
We test three RGB-T semantic segmentation algorithms achieving state-of-the-art results in urban RGB-T scene segmentation benchmarks: CRM \cite{shin2023complementary}, EAEFNet~\cite{eaefnet}, and CMNeXt~\cite{zhang2023delivering}. To compare against single-modality models, we train two additional ResNet50/DeepLabV3+ segmentation models on RGB and thermal data accordingly (\cref{sec:thermal-sem-seg}). 

\subsubsection{Performance analysis}
All RGB-T models outperform the single-modality baselines, with CMNeXt outperforming the next best RGB-T model by 0.026 mIoU (\cref{tab:rgbt-supervised-baselines}). Overall, incorporating RGB imagery into the segmentation pipeline greatly improved the distinction between land-based \textit{stuff} classes which can be hard to distinguish in thermal due to thermal crossover. However, these improvements come at a high computational cost and would not be suitable for field robotic applications. Excelling in this benchmark would involve improving the segmentation performance of the \textit{object} classes, possibly by leveraging RGB data in unsupervised domain adaptation training schemes.

\subsection{RGB-Thermal Image Translation}
We use the RGB-T paired dataset from \cref{sec:rgbt-sem-seg} to benchmark image translation algorithms operating in the RGB$\rightarrow$T direction.  

\subsubsection{Baselines}
We benchmark GAN methods that need RGB-T pairs (Pix2Pix~\cite{isola2017image}, Pix2PixHD~\cite{wang2018pix2pixHD}, VQGAN~\cite{esser2021taming}) and unpaired methods (UNIT~\cite{unit}, MUNIT~\cite{munit}, \cite{lee2023edge}) that do not. We also evaluate Palette~\cite{saharia2022palette}, a paired diffusion approach.

\begin{figure}[htbp]
    \centering
    \includegraphics[width=\figwidth\linewidth]{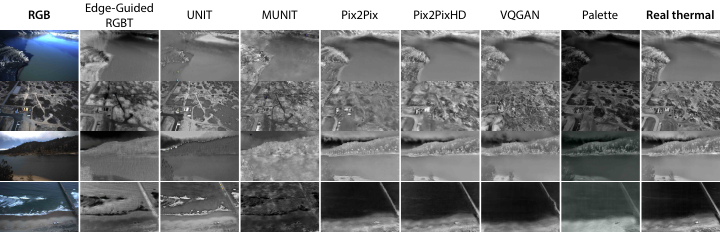}
    \caption{RGB-to-thermal image translation results. Zoom-in to see fine-details.}
    \label{fig:rgbt-translation}
\end{figure}
\subsubsection{Image translation metrics}
We quantify RGB-T image translation using the Peak-Signal-to-Noise ratio (PSNR) and the Structural Similarity Index (SSIM). In addition, we propose a third metric: the thermal mIoU. This is equivalent to the FCN-score used in RGB image translation works~\cite{isola2017image}. Instead of an RGB network, our metric uses our EfficientViT thermal segmentation network (\cref{tab:supervised-baselines}) to segment translated thermal images before evaluating with ground truth. 

\begin{table}[tbp]
	\setlength{\aboverulesep}{1pt}
	\setlength{\belowrulesep}{1pt}
	\centering
	\caption{RGB-Thermal image translation results}
	\label{tab:rgbt-image-translation}
	\resizebox{\linewidth}{!}{%
		\begin{tabular}{p{5.5cm}p{5cm}P{2cm}P{2cm}P{2cm}}
			\toprule[1pt]
			\multirow{2}{*}[-2pt]{Method} & \multirow{2}{*}[-2pt]{Method Type} & \multicolumn{3}{c}{RGB $\rightarrow$ Thermal }\\
			\cmidrule(lr){3-5}
			                                     &                  & PSNR $\uparrow$ & SSIM $\uparrow$ & mIoU $\uparrow$\\
			\midrule[1pt]
			UNIT~\cite{unit}                     & Unpaired GAN     & 12.85                              & 0.421                              & 0.189\\
			MUNIT~\cite{munit}                   & Unpaired GAN     & 16.08                              & 0.459                              & 0.173\\
			Edge-guided RGB-T~\cite{lee2023edge} & Unpaired GAN     & 12.20                              & 0.436                              & 0.225\\
			Pix2Pix~\cite{isola2017image}        & Paired GAN       & \textbf{20.06}                              & 0.547                              & 0.354\\
			Pix2PixHD~\cite{wang2018pix2pixHD}   & Paired GAN       & 20.05                              & \textbf{0.579}                              & 0.379\\
			VQGAN~\cite{esser2021taming}        & Paired GAN      & 18.09  & 0.540        &  \textbf{0.388} \\
                Palette~\cite{saharia2022palette}    & Paired Diffusion & 11.51                              & 0.399                              & 0.194\\
			\bottomrule[1pt]
		\end{tabular}
	}
\end{table}
\subsubsection{Performance analysis}
Paired GANs outperform unpaired GANs and diffusion methods, achieving higher values across all three metrics (\cref{tab:rgbt-image-translation}). Qualitative assessment (\cref{fig:rgbt-translation}) shows poor translations from unpaired techniques, with most retaining geometric artifacts unique to RGB (ocean waves and shadows) and ignoring relative temperature characteristics. Paired GANs produce results similar to real thermal images but with inconsistent details upon closer inspection. Likewise, the diffusion method produces accurate translations but is not consistent. Overall, RGB-T translation still requires improvements before being able to provide a reliable source of thermal training data from existing RGB datasets. For domain adaptation purposes, we emphasize that maximizing the thermal mIoU score is more important than photometric consistency.

\subsection{Motion Tracking}
To quantify VIO/SLAM robustness, we select 12 clipped sequences from our dataset ranging in motion tracking difficulty: from urban sports fields (easy) to our natural environments (hard). We normalize thermal images using the 5\textsuperscript{th} and 95\textsuperscript{th} percentile pixel values (\cref{section:thermal16bit-normalization}) to enable feature detection.

\begin{table}[htbp]
	\setlength{\aboverulesep}{1pt}
	\setlength{\belowrulesep}{1pt}
	\centering
	\caption{VIO/SLAM performance (Absolute Trajectory Error [m]) on aerial sequences}
	\label{table:motiontracking}
	\resizebox{\linewidth}{!}{%
		\begin{tabular}
			{p{2.5cm}p{1.5cm}p{1.75cm}P{1.0cm}P{1.0cm}P{1.0cm}P{1.0cm}P{1.0cm}P{1.0cm}P{1.0cm}P{1.0cm}P{1.0cm}P{1.0cm}P{1.0cm}P{1.0cm}}
			\toprule[1.5pt]

			\multirow{2}{*}[-2pt]{Method} & \multirow{2}{*}[-2pt]{Type} & \multirow{2}{*}[-2pt]{Modality} & \multicolumn{6}{c}{North Field} & \multicolumn{4}{c}{Castaic Lake, CA} & \multicolumn{2}{c}{Duck, NC}\\ \cmidrule(lr){4-9} \cmidrule(lr){10-13} \cmidrule(lr){14-15}
			                   &      &   & N1    & N2    & N3    & N4    & N5    & N6    & C1    & C2    & C3    & C4    & D1    & D2\\
			\midrule[1.5pt]
                VINS-Fusion & \multirow{2}{*}{SLAM} & RGB     & 2.883 & 7.530 & 4.036 & 2.009 & 4.045 & 0.834 & 1.555 & ---   & 3.277 & 1.566 & 1.031 & 0.700\\
			VINS-Fusion &  & Thermal & 9.454 & 12.26 & 9.854 & 2.926 & 11.00 & 5.114 & 7.296 & 1.513 & 5.321 & 2.025 & 0.879 & ---\\
                \midrule[0.25pt]
			VINS-Fusion & \multirow{2}{*}{VIO}  & RGB     & 5.458 & 9.509 & 6.518 & 2.121 & 6.026 & 1.277 & 1.555 & ---   & 3.277 & 1.377 & 1.031 & 0.725\\
			VINS-Fusion &   & Thermal & 13.82 & 12.86 & 11.99 & 2.926 & 13.98 & 5.194 & 7.284 & 10.60 & 5.321 & 2.034 & 0.879 & ---\\
                \midrule[0.25pt]
                Open-VINS & \multirow{2}{*}{VIO}   & RGB     & 22.80 & 36.15 & 37.15 & 5.165 & ---   & 3.896 & ---   & 3.573 & ---   & ---   & 0.475 & 0.700\\
			Open-VINS &     & Thermal & 14.43 & 30.64 & 16.12 & 2.258 & ---   & 1.562 & ---   & 5.311 & ---   & ---   & 1.073 & ---\\
			\midrule[1pt]
                \multicolumn{3}{c}{Trajectory length (m)} & 1533 & 1326 & 1253 & 153 & 1104 & 642 & 310 & 65 & 174 & 137 & 88 & 80 \\
			\bottomrule[1.5pt]
		\end{tabular}
	}%
\end{table}

\subsubsection{Baselines}
We evaluate VINS-Fusion~\cite{qin2018vins} (graph optimization-based) and Open-VINS~\cite{geneva2020openvins} (filtering-based). We test VINS-Fusion in VIO and SLAM modes, with and without loop closure constraints, respectively, and report Absolute Trajectory Error~\cite{sturm2012benchmark} averaged over four runs. 

\subsubsection{Performance analysis}

This benchmark reveals challenges due to fast motion at altitude and periodic feature sparsity in littoral environments. In such scenes, motion tracking frequently failed due to loss of features, requiring sequences to be clipped for quantifiable evaluations (\cref{table:motiontracking}).  VINS-Fusion outperformed Open-VINS mainly due to higher feature tracking reliability. VINS-Fusion works better on RGB images than on thermal images in general. However, VINS-Fusion demonstrates superior feature tracking on thermal images compared to RGB on the C2 sequence of Castaic Lake, where the thermal contrast of stone grains on the lake coast was particularly pronounced in the late afternoon. This observation motivates future research on integrating thermal and RGB cameras for motion tracking with enhanced robustness and versatility. Our benchmark presents a difficult challenge for VIO and SLAM algorithms in extreme environments where water and reflections off water surfaces dominate image scenes, providing a unique test bed to invigorate interest in robust feature matching and motion tracking in natural texture-deficient scenarios.

\subsection{Further Analysis}
\subsubsection{Zero-shot foundation models on thermal imagery}
Current foundation models work well on RGB imagery but have not been benchmarked on thermal, especially in non-urban settings. To evaluate these models on thermal imagery, specifically for zero-shot segmentation and semantic segmentation, we compare RGB and thermal performance using the aligned set (\cref{sec:dataset-rgbt-alignment}) in order to isolate modality as the root cause for performance differences. In the following tests, our metrics do not include the rare \textit{object} classes (\textit{vehicle}, \textit{person}) as they drop metrics to near-uniform values for both modalities, making it difficult to compare.  
\begin{figure}[htb]
    \centering
    \includegraphics[width=\figwidth\linewidth]{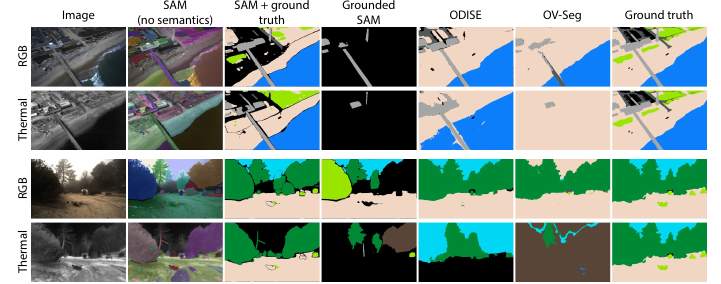}
    \caption{Zero-shot foundation models on RGB-T pairs. Oversegmented SAM outputs differ across modalities but are similar with ground truth semantics added. Semantic segmentation models (prompted with all classes) perform better on RGB imagery.}
    \label{fig:zeroshot-segmentation-results}
\end{figure}

\begin{table}[tb]
    \caption{RGB foundation model performances on thermal imagery}
    \vspace{-0.25cm}
    \begin{subtable}[t]{0.49\linewidth}
        \caption{\textbf{Instance} segmentation via SAM on thermal and augmented color images using coregistered color image segmentations as ground truth}
        \resizebox{\linewidth}{!}{%
        \begin{tabular}{p{2.5cm}P{1.2cm}P{1.2cm}P{1.2cm}P{1.2cm}}
            \toprule[1pt]
            \textbf{Modality} & AP$^\dagger_{\textrm{all}}$ & AP$_{\textrm{small}}$ & AP$_{\textrm{med}}$ & AP$_{\textrm{large}}$  \\
            \midrule[1pt]
            Thermal & 0.018 & 0.007 & 0.022 & 0.204 \\
            Color (grayscale) & 0.538 & 0.536 & 0.531 & 0.601 \\
            Color (col. jitter) & 0.800 & 0.792 & 0.801 & 0.841 \\
            \bottomrule[1pt]
        \end{tabular}
        }
        \caption*{$^\dagger$refers to AP@0.5::0.95, following MSCOCO~\cite{lin2014microsoft}}
        \label{tab:sam-map}
    \end{subtable}
    \hfill
    \begin{subtable}[t]{0.49\linewidth}
        \caption{Zero-shot \textbf{semantic} segmentation (mIoU) on coregistered color and thermal imagery}
        \vspace{0.12cm}
        \resizebox{\linewidth}{!}{%
        \begin{tabular}{p{3.25cm}P{1.5cm}P{1.5cm}P{1.5cm}P{1.5cm}}
            \toprule[1pt]
            \textbf{Method} & \textbf{Color} & \bfseries\makecell{Color \\ (gray)} & \bfseries\makecell{Color \\ (jitter)} & \textbf{Thermal} \\
            \midrule[1pt]
            \makecell[l]{SAM~\cite{kirillov2023segany} + GT ann.$^{\ddagger}$} & 0.704 & 0.698 & 0.700 & 0.653 \\
            \makecell[l]{Grounded SAM~\cite{liu2023grounding}} & 0.380 & 0.351 & 0.361 & 0.193 \\
            \makecell[l]{OV-Seg~\cite{liang2023open}} & 0.362 & 0.340 & 0.365 & 0.217 \\
            \makecell[l]{ODISE~\cite{xu2023open}} & 0.504 & 0.450 & 0.486 & 0.232 \\
            \bottomrule[1pt]
        \end{tabular}
        }
        \caption*{$^\ddagger$ Most frequent ground truth class label in a SAM mask assigned as label for entire mask}
        \label{tab:zero-shot-miou-small}
    \end{subtable}    
    \label{tab:foundation}
\end{table}

We quantify Segment Anything's (SAM)~\cite{kirillov2023segany} zero-shot segmentation sensitivity to modality change by measuring the spatial alignment of its mask outputs. We apply it to the RGB set and the 8-bit thermal image set (\cref{section:thermal16bit-normalization}) with the default 32$\times$32 grid points to segment everything (\cref{fig:zeroshot-segmentation-results}). We compute the average precision (AP) of the predicted thermal masks, using the predicted RGB masks as ground truth, and compare the thermal AP to the APs of a color-jittered RGB set and a grayscale set. We find that SAM is sensitive to photometric changes, with large performance drops from color-jitter to grayscale to thermal (\cref{tab:sam-map}).

Next, we test large vision-language models (Grounded-SAM~\cite{liu2023grounding}, OV-Seg~\cite{liang2023open}, and ODISE~\cite{xu2023open}) in zero-shot semantic segmentation. We prompt the models with a list of classes in our dataset; generate semantic segmentation masks for RGB, color-jittered RGB, grayscale, and thermal image sets; and compute the mIoU of their outputs using our ground truth annotations. Although RGB, RGB-variants, and thermal modalities all yield poor semantic segmentation performance (\cref{tab:zero-shot-miou-small}), all RGB variants outperform thermal by at least 0.13 mIoU, indicating low generalization to the thermal modality (\cref{fig:zeroshot-segmentation-results}). 

Finally, we create semantic SAM masks by assigning class labels to SAM instances based on the most frequent ground truth class within an instance (\cref{fig:zeroshot-segmentation-results}). We find that these masks attain much higher mIoU score (0.653) compared to the other zero-shot methods. Based on the results (\cref{tab:foundation}), we note that SAM can find object boundaries in the thermal domain reasonably well but can oversegment, resulting in poor AP scores when compared to RGB (which can also oversegment). When taken with ground truth semantics, this behavior is hidden.  

\subsubsection{Transfer learning with urban thermal datasets}
\begin{wraptable}{r}{0.5\linewidth}
\setlength{\aboverulesep}{0pt}
    \setlength{\belowrulesep}{0pt}
    \centering
    \vspace{-0.25cm}
    \caption{Network pretraining methods for downstream thermal segmentation.}
    \label{tab:ablation-thermal-urban}
    \resizebox{\linewidth}{!}{%
    \begin{tabular}{p{2.5cm}P{1.5cm}P{1.5cm}P{2.5cm}P{2.5cm}}
        \hlineB{3}
        \multirow{2}{*}{Model} & \multicolumn{3}{c}{mIoU} & \multirow{2}{*}{\makecell{\#\\ Params. (M)}}\\ \cmidrule{2-4} 
        & None & ImageNet & Therm. Urban & \\ 
        \hlineB{3}
        FastSCNN & 0.640 & 0.690 & 0.688 & 1.1 \\ 
        EfficientViT & 0.687 & 0.725 & 0.714 & 4.8 \\ 
        ResNet18 & 0.702 & 0.713 & 0.706 & 12.3 \\ 
        ResNet50 & 0.682 & 0.713 & 0.728 & 26.7 \\ 
        ConvNext-B & 0.625 & 0.717 & 0.697 & 89.4 \\
        \hlineB{3}
    \end{tabular}
    }
\end{wraptable}

We test if pretraining on thermal urban data can assist transfer to field settings. We sample 40k unlabeled thermal images from \cite{flir, vertens2020heatnet, suo2022hit, choi2018kaist, liu2022target, ha2017mfnet, shin2023deep, XIONG2021103628} and pretrain various networks using MoCoV2~\cite{chen2020mocov2}. We then finetune the pretrained networks on our training set and evaluate on our test set. Overall, we find sparse evidence suggesting any advantage of pretraining with existing urban thermal datasets (\cref{tab:ablation-thermal-urban}). Off-the-shelf ImageNet weights can provide strong performance in the thermal field domain, especially over training from scratch, without the effort and costs of pretraining.

%% file: sections/conclusion.tex
\section{Conclusion}
\label{sec:conclusion}
We presented the Caltech Aerial RGB-Thermal Dataset, the first dataset tailored towards advancing thermal semantic perception and motion tracking algorithms in natural environments. We created four benchmarks encompassing semantic segmentation, image translation, and motion tracking, and demonstrate challenges faced by current thermal perception and localization algorithms. Semantic segmentation and image translation methods were notably affected by geographical domain shifts, reflecting the wide intra-class and inter-scenery distributions in our dataset, and temporal domain shifts, due to thermal inversion and crossover. 
\rebuttal{Motion tracking in thermal suffered from feature sparsity, a challenge commonly encountered in real-world robotic deployments}~\cite{Ackerman_2024}. Our motion tracking benchmark penalizes current VIO/SLAM algorithms for assuming ideal feature-tracking conditions, serving as a testbed to fuel improvements in this area. Our dataset and benchmarks can be used into the future to help develop algorithms that expand the operating domains of robotics and computer vision.

%% file: supplemental_material.tex
\section{Comparison of Related Benchmarks and Datasets}
\label{sec:sup-related-work}
A detailed comparison of related datasets and benchmarks is shown in \cref{table:sup-related-datasets}. Included datasets must be captured from aerial vehicles, surface vehicles (boats), or cars. We included pure RGB datasets only if they are related to semantic segmentation or object detection and captured from aerial vehicles. Other datasets included for comparison must contain thermal imagery and be related to semantic segmentation or object detection. 

\begin{table*}[htbp]
    \setlength{\tabcolsep}{10pt}
    \centering
    \caption{Comparison of datasets captured from aerial platforms or depicting thermal scenes.}
    \label{table:sup-related-datasets}
    \resizebox{\linewidth}{!}{
    \begin{tabular}{lllccccllcc}
        \toprule[1pt]
        Dataset & Platform & Task & RGB & Thermal & GPS & IMU & Setting & Location & \# Samples & \makecell{Camera pose} \\
        \midrule[1pt]

        \textbf{Ours} & UAV & Sem. Seg. & \customcheck & \customcheck & \customcheck & \customcheck & \makecell[l]{River, Coast, Lake, \\Mountain, Desert} & USA & 4,195 & 20\degree, 45\degree, < 120 m \\
        \midrule[0.25pt]

        \makecell[l]{AeroScapes~\cite{nigam2018ensemble}, \\ UDD~\cite{chen2018large}, UAVid~\cite{LYU2020108},\\ VDD~\cite{cai2023vdd}, IDD~\cite{cai2023vdd}} 
        & UAV & Sem. Seg. & \customcheck & \customcross & \customcross  & \customcross & Urban & China & 205 - 3269 & 30 - 90\degree, < 100 m \\
        \midrule[0.25pt]
        
        \makecell[l]{Semantic Drone \\ Dataset~\cite{semanticDroneDataset}} & UAV & Sem. Seg. & \customcheck & \customcheck & \customcross & \customcheck & Urban & Germany & 400 & 90\degree, < 30m \\
        \midrule[0.25pt]

        \makecell[l]{Swiss and Okutama \\Drone Datasets~\cite{speth2022deep}} & UAV & Sem. Seg. & \customcheck & \customcross  & \customcross  & \customcross & Urban & \makecell[l]{Switzerland, \\Japan} & 191 & 90\degree \\
        \midrule[0.25pt]
        
        \makecell[l]{NII‐CU Multispectral \\Aerial Person Detection~\cite{speth2022deep}} & UAV & Person Det. & \customcheck & \customcheck  & \customcross  & \customcross & Urban & Japan & 5,880 & 45\degree \\
        \midrule[0.25pt]

        WIT-UAS~\cite{jong2023wit} & UAV & Obj. Det. & \customcross & \customcheck & \customcheck & \customcheck & \makecell[l]{Forest} & USA & 6,951 & 30\degree, 90\degree, < 120 m \\
        \midrule[0.25pt]
        
        VisDrone~\cite{zhu2021detection} & UAV & Obj. Det. & \customcheck & \customcross  & \customcross  & \customcross & Urban & China & 10,209  & Variable \\
        \midrule[0.25pt]
        MassMIND~\cite{nirgudkar2023massmind} & USV & Sem. Seg. & \customcross & \customcheck & \customcheck & \customcheck & Harbor & USA & 2,900 & 0\degree, 0 m \\
        \midrule[0.25pt]
        Flir ADAS~\cite{flir} & Car & Obj. Det. & \customcheck & \customcheck  & \customcross & \customcross & Urban & USA & 10,228 & 0\degree, 0 m \\
        \midrule[0.25pt]
        BIRDSAI~\cite{bondi2020birdsai} & UAV & Obj. Track. & \customcross & \customcheck & \customcross & \customcross & Savannas & S. Africa & 62,000 & Off-nadir, 60-120 m \\
        \midrule[0.25pt]
        HIT-UAV~\cite{suo2022hit} & UAV & Obj. Det. & \customcheck & \customcheck & \customcross & \customcross & Urban & China & 2,898 & 30\degree-90\degree, 60-130 m \\
        \midrule[0.25pt]
        \makecell[l]{KAIST \\ Multispectral~\cite{choi2018kaist}} & Car & Obj. Det. & \customcheck & \customcheck & \customcheck & \customcheck & Urban & S. Korea & 4750 & 0\degree, 0 m \\ 
        \midrule[0.25pt]
        MFNet~\cite{ha2017mfnet} & Car & Sem. Seg. & \customcheck & \customcheck & \customcross & \customcross & Urban & Japan & 1,569 & 0\degree, 0 m \\
        \midrule[0.25pt]
        M$^3$FD~\cite{liu2022target} & Car & Obj. Det. & \customcheck & \customcheck & \customcross & \customcross & Urban & China & 4,200 & 0\degree, 0 m \\
        \midrule[0.25pt]

        Freiburg Thermal~\cite{vertens2020heatnet} & Car & Sem. Seg. & \customcheck & \customcheck & \customcross & \customcross & Urban & Germany & 20,656$^\ddagger$ & 0\degree, 0 m \\
        \midrule[0.25pt]

        SODA~\cite{li2020segmenting}, SCUT-Seg~\cite{XIONG2021103628} & Car & Sem. Seg. & \customcross & \customcheck & \customcross & \customcross & Indoor/Urban & --- & $\sim$2,000 & 0\degree, 0 m \\
        \midrule[0.25pt]

        STheReO~\cite{yun2022sthereo}, MS$^2$~\cite{shin2023deep} & Car & \makecell[l]{SLAM / \\Depth Est.} & \customcheck & \customcheck & \customcheck & \customcheck & \makecell[l]{Urban/Suburban} & S. Korea & --- & 0\degree, 0 m \\
        \midrule[0.25pt]

        PST900~\cite{shivakumar2020pst900} & UGV & Sem. Seg. & \customcheck & \customcheck & \customcross & \customcross & \makecell[l]{Subterranean} & USA & 894 & 0\degree, 0 m \\
        \midrule[0.25pt]
        
        LLVIP~\cite{jia2021llvip} & \makecell[l]{Building \\ (fixed)} & \makecell[l]{Pedestrian\\ Detection} & \customcheck & \customcheck & \customcross & \customcross & Urban & China & 14,588 & Variable \\        

        \bottomrule[1pt]
    \end{tabular}
    }
    
    \captionsetup{font=scriptsize}
    \caption*{$^\dagger$ Camera angle of 90\degree is nadir-pointing \quad $^\ddagger$ Annotations provided for only 64 test set images.}
\end{table*}

\section{Dataset Information}
\label{sec:sup-dataset-info}
\begin{figure}[htbp]
    \centering
    \includegraphics[width=0.95\linewidth]{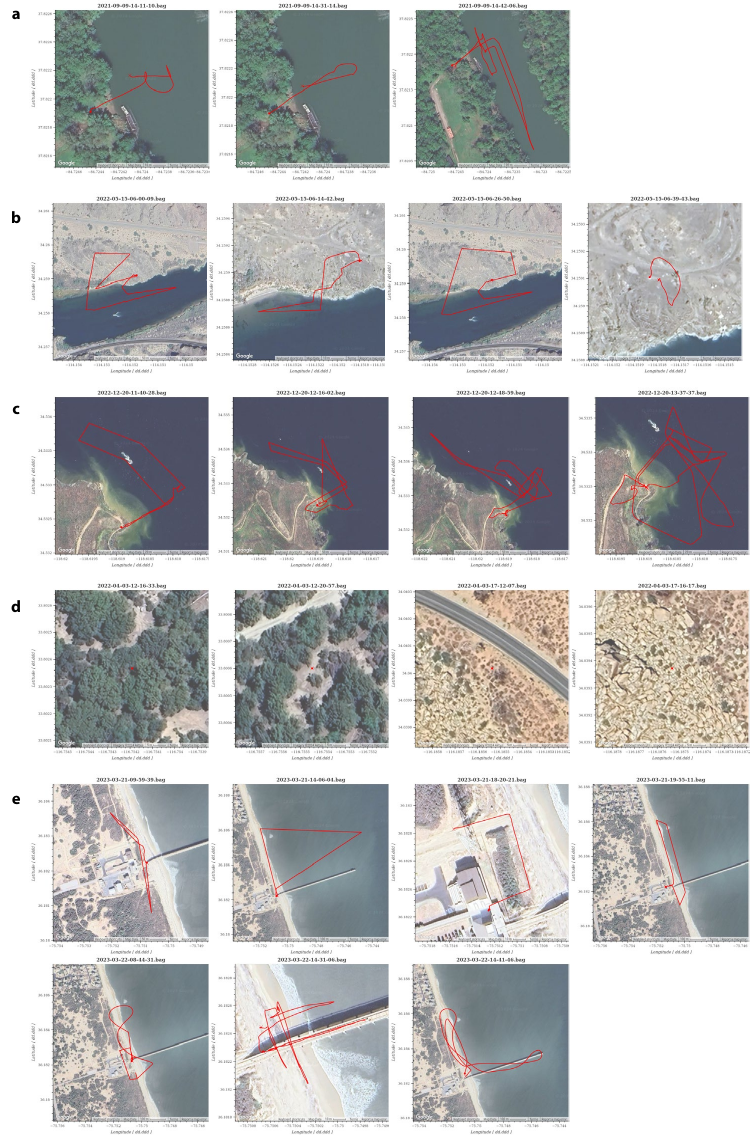}
    \caption{Trajectories from our dataset: \textbf{(a)} Kentucky River \textbf{(b)} Colorado River \textbf{(c) }Castaic Lake \textbf{(d)} Idyllwild and Joshua Tree \textbf{(e) }Duck. North Field trajectories not shown.}
    \label{fig:sup-flight-trajectories}
\end{figure}

Our dataset consists of 37 aerial and ground trajectories captured from diverse natural landscapes across the USA. Further details are shown in \cref{table:sup-datasets}. Visualizations of several flight trajectories are shown in \cref{fig:sup-flight-trajectories}.
\begin{table}[h]
    \centering
    \caption{Dataset capture locations and settings.}
    \label{table:sup-datasets}
    \resizebox{\linewidth}{!}{%
    \begin{tabular}{p{4cm}p{2.3cm}p{2.8cm}p{2cm}p{2cm}>{\centering\arraybackslash}p{1.3cm}>{\centering\arraybackslash}p{2.0cm}}
    \toprule[1pt]
    Location of Capture & Terrain Type & Capture Method & Motion & Time of Day & \# Seq. & Total Time \\
    \midrule[1pt]
    Kentucky River, KY  & River     & UAV Flight    & Large    & Afternoon   & 3 & 17m10s \\
    Colorado River, CA  & River     & UAV Flight    & Large  & Sunrise     & 4 & 32m50s \\
    Castaic Lake, CA    & Lake      & UAV Flight    & Large  & Midday      & 4 & 58m56s \\
    Duck, NC            & Coast     & UAV Flight    & Large  & Day/Night   & 7 & 91m26s \\
    Big Bear Lake, CA   & Lake      & Ground        & Minor  & Mid-morning & 8 & 15m52s \\
    Arroyo Seco, CA     & Stream    & Ground        & Still  & Afternoon   & 8 & 10m22s \\
    Idyllwild, CA       & Mountain  & Ground        & Minor  & Day         & 2 &  5m58s \\
    Joshua Tree, CA     & Desert    & Ground        & Minor  & Day         & 2 &  5m58s \\
    
    North Field (Caltech), CA     & Urban     & UAV Flight    & Large  & Day         & 5 & 26m13s \\
    \bottomrule[1pt]
    \end{tabular}
    }
\end{table}

\subsection{Data Capture}
\label{sec:sup-dataset-capture-info}
All data is stored as ROS1 rosbags as this is a natural format for robotics work.
As the rosbag format may not be preferred for all users, extraction tools for generating csv and images files are provided at \textit{link\_provided\_after\_review}.
Within the rosbags, the synchronized data contains timestamps for both the time of trigger (\textit{sync/rate\_*}) and time the data was received (header stamp in the topic), allowing the data to be aligned in post-processing regardless of transport delays in the system.
Where available, position data should be taken from the UAV (\textit{uav1/mavros/local\_position/*} topics) and orientation data from the VN100 (\textit{imu/imu}) to provide the best available estimate.
Finally, while mostly complete, not all datasets contain all sensors due to issues when collecting data. 

\subsection{Sensor Calibration}
\label{sec:sup-sensor-calibration-info}
We calibrate the three cameras and IMU via Kalibr~\cite{kalibr} by conducting three independent calibrations. We do this to isolate any difficulties caused by the thermal camera. We first calibrate the thermal camera and IMU, and then perform stereo calibration for each possible camera pairing. Thermal calibrations are done using a 10$\times$10 circle grid (1$''$ diameter). To ensure calibration image sharpness in the thermal domain, we place the calibration board in direct sunlight for 2-3 minutes prior to data collection and keep it illuminated (reflecting sunlight) during the entire collection process.
The RGB and Mono cameras were calibrated using a standard 6x6 April tag grid board.

\subsection{Labeling Process for Semantic Segmentation Annotations}
\label{sec:sup-dataset-semseg-annotation}
The thermal images were labeled via manual annotation. Annotators were asked to read a guide with descriptions of each trajectory. Descriptions contained lists of possible classes, a expert-labeled examples provided by the authors, and GPS coordinates (for look-up in Google Earth). During labeling, annotators were presented with a side-by-side copy of a thermal image and its corresponding RGB image, the trajectory it came from, and were asked to annotate the thermal image. The annotations were reviewed by authors, with rejected annotations sent back for re-annotation. A total of 3 rounds of annotation were conducted.

\section{Implementation and Training Details} 
\label{sec:sup-impl-info}
Our baselines were implemented using code from public Github repositories shown in \cref{tab:sup-implementation}. Official code was used whenever possible. We trained and tested all networks using a single Nvidia A6000 ADA GPU (48 GB).

\begin{table}[h]
    \setlength{\tabcolsep}{10pt}
    \centering
    \caption{Public repositories used in this work to create our baselines}
    \resizebox{\linewidth}{!}{%
    \begin{tabular}{p{3.5cm}p{13cm}}
        \toprule[1.5pt]
        Baselines & Public Repositories \\
        \midrule[1.5pt]
        \multicolumn{2}{l}{\textbf{Thermal Segmentation}} \\
        \midrule[1pt]
        FastSCNN~\cite{poudel2019fast} & \path{Tramac/Fast-SCNN-pytorch} \\
        EfficientViT~\cite{cai2022efficientvit} & \path{mit-han-lab/efficientvit} \\
        Segformer~\cite{xie2021segformer} & \path{NVlabs/SegFormer} \\
        DINOv2~\cite{oquab2023dinov2} & \path{facebookresearch/dinov2} \\
        FTNet~\cite{ftnet} & \path{shreyaskamathkm/FTNet} \\
        Everything else \cite{howard2019searching, mehta2022separable, tan2019efficientnet, he2016deep, xie2017aggregated, liu2022convnet} & \mbox{\path{huggingface/pytorch-image-models} +} \mbox{\path{qubvel/segmentation_models.pytorch}} \\
        \midrule[1pt]
        \multicolumn{2}{l}{\textbf{RGB-T Segmentation}} \\
        \midrule[1pt]
        EAEFNet~\cite{eaefnet}            & \path{FreeformRobotics/EAEFNet} \\
        CRM~\cite{shin2023complementary} & \path{UkcheolShin/CRM_RGBTSeg}  \\
        CMNeXt~\cite{zhang2023delivering} & \path{jamycheung/DELIVER} \\
        \midrule[1pt]
        \multicolumn{2}{l}{\textbf{RGB-T Image Translation}} \\
        \midrule[1pt]
        UNIT~\cite{unit}                     & \path{NVlabs/imaginaire}   \\
        MUNIT~\cite{munit}                    & \path{NVlabs/imaginaire}   \\
        Edge-guided RGB-T~\cite{lee2023edge} & \path{RPM-Robotics-Lab/sRGB-TIR}   \\
        Pix2Pix~\cite{isola2017image}         & \path{junyanz/pytorch-CycleGAN-and-pix2pix}  \\
        Pix2PixHD~\cite{wang2018pix2pixHD}    & \path{NVIDIA/pix2pixHD}  \\
        VQ-GAN~\cite{esser2021taming}         & \path{CompVis/taming-transformers}  \\
        Palette~\cite{saharia2022palette}    & \path{Janspiry/Palette-Image-to-Image-Diffusion-Models}  \\
        \midrule[1pt]
        \multicolumn{2}{l}{\textbf{VIO/SLAM}} \\
        \midrule[1pt]
        VINS-Fusion~\cite{qin2018vins}                     & \path{HKUST-Aerial-Robotics/VINS-Fusion}  \\
        OpenVINS~\cite{geneva2020openvins}                    & \path{rpng/open_vins}  \\
        \bottomrule[1.5pt]
    \end{tabular}
    }
    
    \label{tab:sup-implementation}
\end{table}

\subsection{Thermal Baselines}
\label{sec:sup-thermal-impl-info}
Most networks used in our thermal baseline experiments (Tab.~2) employed a DeepLabV3+ segmentation head with the exception of FastSCNN~\cite{poudel2019fast}, EfficientViT~\cite{cai2022efficientvit}, Segformer~\cite{xie2021segformer}, DINOv2, and FTNet. For DINOv2, we implemented the linear and nonlinear multi-scale segmentation heads ourselves, with their architectures shown in \cref{tab:sup-dinov2-seg}. All networks, besides DINOv2 and ConvNeXt (CLIP), were trained starting from pretrained ImageNet weights.    
\begin{table}[h]
    \centering
    \caption{Multi-scale segmentation head architecture used with DINOv2. The architecture is described using PyTorch syntax~\cite{paszke2019pytorch}.}
    \label{tab:sup-dinov2-seg}
    \resizebox{\linewidth}{!}{%
    \begin{tabular}{P{7cm}|P{7cm}}
        \hlineB{3}
         DINOv2 Linear Head &  DINOv2 Nonlinear Head  \\
         \hlineB{3}
         \texttt{nn.Conv2d(384*4, 10, 1)} & \texttt{nn.Conv2d(384*4, 512, 3, padding=1)} \\
         \hlineB{1}
            & \texttt{nn.GELU} \\
        \hlineB{1}
            & \texttt{nn.Conv2d(512, 10, 1)} \\
         \hlineB{3}
    \end{tabular}
    }
\end{table}

Networks were trained for 300 epochs or until validation loss plateued. We used the Adam optimizer with a \texttt{1e-3} learning rate that decayed at an exponential rate of \texttt{0.99} and a batch size of 64. We augmented our 16-bit thermal training data using random contrast stretches (within the lower and upper 5\textsuperscript{th} percentiles) and CLAHE with a random clip limit (see Sec.~3.2). Photometric augmentations were followed by horizontal flips, rotations (within 10$\degree$), scaling between 1 and 1.5, and random crops to 512 $\times$ 512.

\subsection{RGB-T Segmentation, Image Translation, and VIO/SLAM}
We used the repositories listed in \cref{tab:sup-implementation} and followed their training procedures. However, we increased the batch size to maximize GPU memory usage whenever possible.

\section{Additional Results}
\label{sec:sup-additional-results}
\subsection{Relative Thermal Pixel Intensities Throughout the Day}
\begin{figure}[h]
    \centering
    \includegraphics[width=\linewidth]{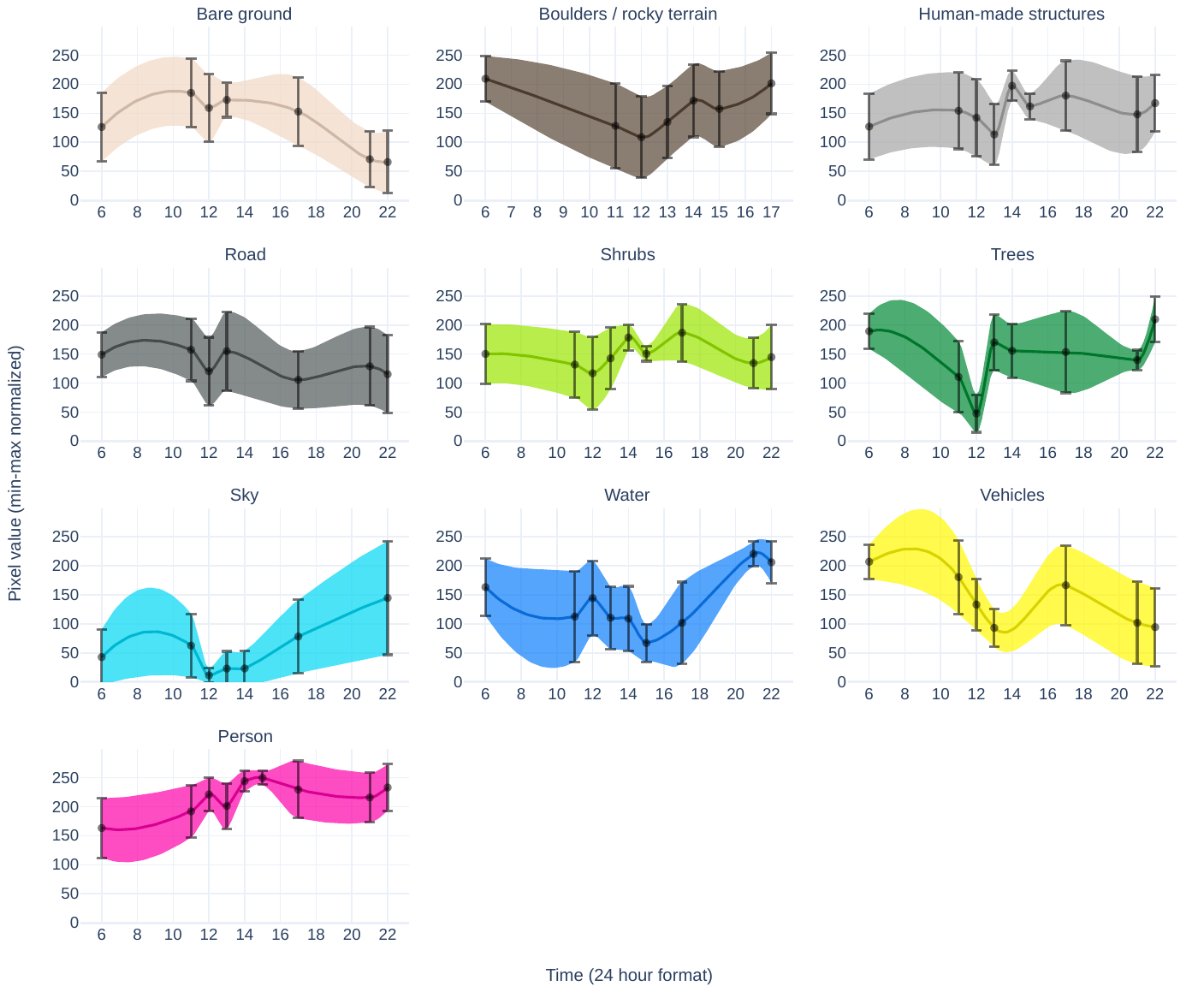}
    \caption{Change in normalized pixel values per class throughout the day.}
    \label{fig:sup-class-pixel-values-by-hour}
\end{figure}

In order to better understand how different objects and entities appear in thermal imagery at different times of day, we plot the thermal pixel intensities of each class (using the ground truth annotations) as a function of their recorded capture time (\cref{fig:sup-class-pixel-values-by-hour}). Since our thermal camera was not radiometric, we plotted the relative thermal pixel values post-normalization using the normalization scheme described in Sec.~3.2. The plots show cases of thermal inversion between certain classes, notably \textit{bare ground} and \textit{water}. 

\section{Examples}
\subsection{Thermal Inversion}
\label{sec:sup-additional-results-thermal-inversion}
\begin{figure}[h]
    \centering
    \includegraphics{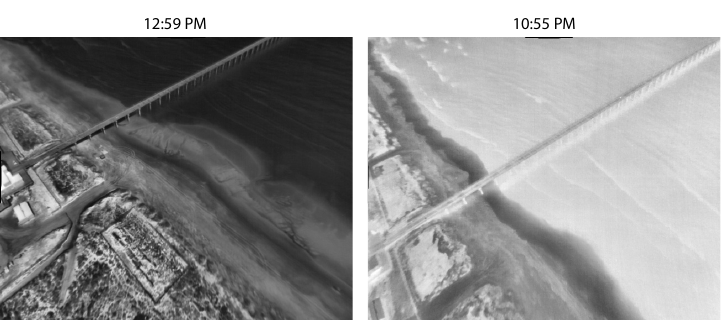}
    \caption{Example of thermal inversion of water and land classes at Duck, NC.}
    \label{fig:sup-thermal-inversion}
\end{figure}

Thermal inversion occurs when object pixel values change depending on its temperature. One notable example of this is thermal inversion of water and land as seen in \cref{fig:sup-thermal-inversion}. In this example, the two images were taken over the same location (Duck, NC) but at different times of day.
 
